\documentclass[sigconf]{acmart}\pdfoutput=1
\makeatletter
\def\mdseries@tt{m}
\makeatother 
\usepackage[draft=true,newfloat]{minted} %2
\usepackage{graphicx}
\usepackage{hyperref}           %5
\usepackage[utf8]{inputenc}
\usepackage{verbatim}
\newsavebox{\mintedbox}
\usepackage{xcolor}
\usepackage{float}
\usepackage{tabularx} 
\usepackage{booktabs}

\usepackage{setspace}
\usepackage{placeins}
\usepackage{tcolorbox}
\usepackage{xparse}
\tcbuselibrary{minted,breakable,xparse,skins}
\usepackage{listings}
\usepackage{caption}
\usepackage{subfig}
\usepackage{pgfplots}

\usepgfplotslibrary{groupplots}
\newenvironment{code}{\captionsetup{type=listing}}{}
\SetupFloatingEnvironment{listing}{name=Listings}

\usepackage{xcolor,pifont}

\definecolor{cycle2}{RGB}{106, 191, 0}
\definecolor{cycle3}{RGB}{191, 0, 0}

\newcommand{\cmark}{\textcolor{cycle2}{\ding{52}}} %
\newcommand{\xmark}{\textcolor{cycle3}{\ding{56}}}

\newcommand{\specialcell}[2][c]{%
	\begin{tabular}[#1]{@{}c@{}}#2\end{tabular}}

\title{PyTorch Geometric Temporal: Spatiotemporal Signal Processing with Neural Machine Learning Models}

\acmConference[CIKM'21]{CIKM'21: ACM International Conference on Information and Knowledge Management}{1-5 November 2021}{Online}
\acmBooktitle{CIKM'21: ACM International Conference on Information and Knowledge Management, 1-5 November 2021, Online}
\begin{document}
\author{Benedek Rozemberczki}
\authornote{The project started when the author was a doctoral student of the Center for Doctoral Training in Data Science at The University of Edinburgh.}
\affiliation{
\institution{AstraZeneca}
  \country{United Kingdom}
}
\email{benedek.rozemberczki@astrazeneca.com}
\author{Paul Scherer}
\affiliation{
  \institution{University of Cambridge}
  \country{United Kingdom}
}
\email{pms69@cam.ac.uk}
\author{Yixuan He}
\affiliation{
  \institution{University of Oxford}
  \country{United Kingdom}
}
\email{yixuan.he@stats.ox.ac.uk}

\author{George Panagopoulos}
\affiliation{
  \institution{École Polytechnique}
  \country{France}
}
\email{george.panagopoulos@polytechnique.edu}

\author{Alexander Riedel}
\affiliation{
  \institution{Ernst-Abbe University for Applied Sciences}
  \country{Germany}
}
\email{alexander.riedel@eah-jena.de}

\author{Maria Astefanoaei}
\affiliation{
  \institution{IT University of Copenhagen}
  \country{Denmark}
}
\email{msia@itu.dk}

\author{Oliver Kiss}
\affiliation{
  \institution{Central European University}
  \country{Hungary}
}
\email{kiss_oliver@phd.ceu.edu}
\author{Ferenc Beres}
\affiliation{
  %\institution{Eotvos University}
  \institution{ELKH SZTAKI}
  \country{Hungary}
}
\email{beres@sztaki.hu}

\author{Guzmán López}
\affiliation{
  \institution{Tryolabs}
  \country{Uruguay}
}
\email{guzman@tryolabs.com}

\author{Nicolas Collignon}
\affiliation{
  \institution{Pedal Me}
  \country{United Kingdom}
}
\email{nicolas@pedalme.co.uk}
\author{Rik Sarkar}
\affiliation{
  \institution{The University of Edinburgh}
  \country{United Kingdom}
}
\email{rsarkar@inf.ed.ac.uk}
\renewcommand{\shortauthors}{B. Rozemberczki, P. Scherer, Y. He, G. Panagopoulos, A. Riedel, M. Astefanoaei, O. Kiss, F. Beres, G. López, N. Collignon, and R. Sarkar}
\begin{abstract}
We present PyTorch Geometric Temporal a deep learning framework combining state-of-the-art machine learning algorithms for neural spatiotemporal signal processing. The main goal of the library is to make temporal geometric deep learning available for researchers and machine learning practitioners in a unified easy-to-use framework. PyTorch Geometric Temporal was created with foundations on existing libraries in the PyTorch eco-system, streamlined neural network layer definitions, temporal snapshot generators for batching, and integrated benchmark datasets. These features are illustrated with a tutorial-like case study. Experiments demonstrate the predictive performance of the models implemented in the library on real world problems such as epidemiological forecasting, ride-hail demand prediction and web-traffic management. Our sensitivity analysis of runtime shows that the framework can potentially operate on web-scale datasets with rich temporal features and spatial structure.
\end{abstract}
\maketitle

\section{Introduction}\label{sec:introduction}
Deep learning on static graph structured data has seen an unprecedented success in various business and scientific application domains. Neural network layers which operate on graph data can serve as building blocks of document labeling, fraud detection, traffic forecasting and cheminformatics systems \cite{chickenpox,rozemberczki2020karate,yu2018spatio,bojchevski2020scaling,rozemberczki2020pathfinder}. This emergence and the wide spread adaptation of geometric deep learning was made possible by open-source machine learning libraries.  The high quality, breadth, user oriented nature and availability of specialized deep learning libraries \cite{pytorch_geometric,dlg,StellarGraph,rozemberczki2020karate} were all contributing factors to the practical success and large-scale deployment of graph machine learning systems. At the same time the existing geometric deep learning frameworks operate on graphs which have a fixed topology and it is also assumed that the node features and labels are static. Besides limiting assumptions about the input data, these off-the-shelf libraries are not designed to operate on spatiotemporal data.

\textbf{Present work.} We propose PyTorch Geometric Temporal, an open-source Python library for spatiotemporal machine learning. We designed PyTorch Geometric Temporal with a simple and consistent API inspired by the software architecture of existing widely used geometric deep learning libraries from the PyTorch ecosystem \cite{pytorch, pytorch_geometric}. Our framework was built by applying simple design principles consistently. The framework reuses existing neural network layers in a modular manner, models have a limited number of public methods and hyperparameters can be inspected. Spatiotemporal signal iterators ingest data memory efficiently in widely used scientific computing formats and return those in a PyTorch compatible format. The design principles in combination with the test coverage, documentation, practical tutorials, continuous integration, package indexing and frequent releases make the framework an end-user friendly spatiotemporal machine learning system. 

The experimental evaluation of the framework entails node level regression tasks on datasets released exclusively with the framework. Specifically, we compare the predictive performance of spatiotemporal graph neural networks on epidemiological forecasting, demand planning, web traffic management and social media interaction prediction tasks. Synthetic experiments show that with the right batching strategy PyTorch Geometric Temporal is highly scalable and benefits from GPU accelerated computing.

\textbf{Our contributions.} The main contributions of our work can be summarized as:
\begin{itemize}
    \item We publicly release \textit{PyTorch Geometric Temporal} the first deep learning library for parametric spatiotemporal machine learning models. 
    \item We provide data loaders and iterators with \textit{PyTorch Geometric Temporal} which can handle spatiotemporal datasets.
    \item We release new spatiotemporal benchmark datasets from the renewable energy production, epidemiological reporting, goods delivery and web traffic forecasting domains. 
    \item We evaluate the spatiotemporal forecasting capabilities of the neural and parametric machine learning models available in \textit{PyTorch Geometric Temporal} on real world datasets. 
\end{itemize}

The remainder of the paper has the following structure. In Section \ref{sec:related_work} we overview important preliminaries and the related work about temporal and geometric deep learning and the characteristics of related open-source machine learning software. The main design principles of \textit{PyTorch Geometric Temporal} are discussed in Section \ref{sec:design} with a practical example. We demonstrate the forecasting capabilities of the framework in Section \ref{sec:experiments} where we also evaluate the scalability of the library on various commodity hardware. We conclude in Section \ref{sec:conclusions} where we summarize the results. The source code of \textit{PyTorch Geometric Temporal} is publicly available at \url{https://github.com/benedekrozemberczki/pytorch_geometric_temporal}; the Python package can be installed via the \textit{Python Package Index}. Detailed documentation is accessible at \url{https://pytorch-geometric-temporal.readthedocs.io/}.

\section{Preliminaries and related work }\label{sec:related_work}
In order to position our contribution and highlight its significance we introduce some important concepts about spatiotemporal data and discuss related literature about geometric deep learning and machine learning software.
\subsection{Temporal Graph Sequences}
\begin{figure}[h!]
\centering
\subfloat[Dynamic graph with temporal signal.]{\includegraphics[height=1.0in]{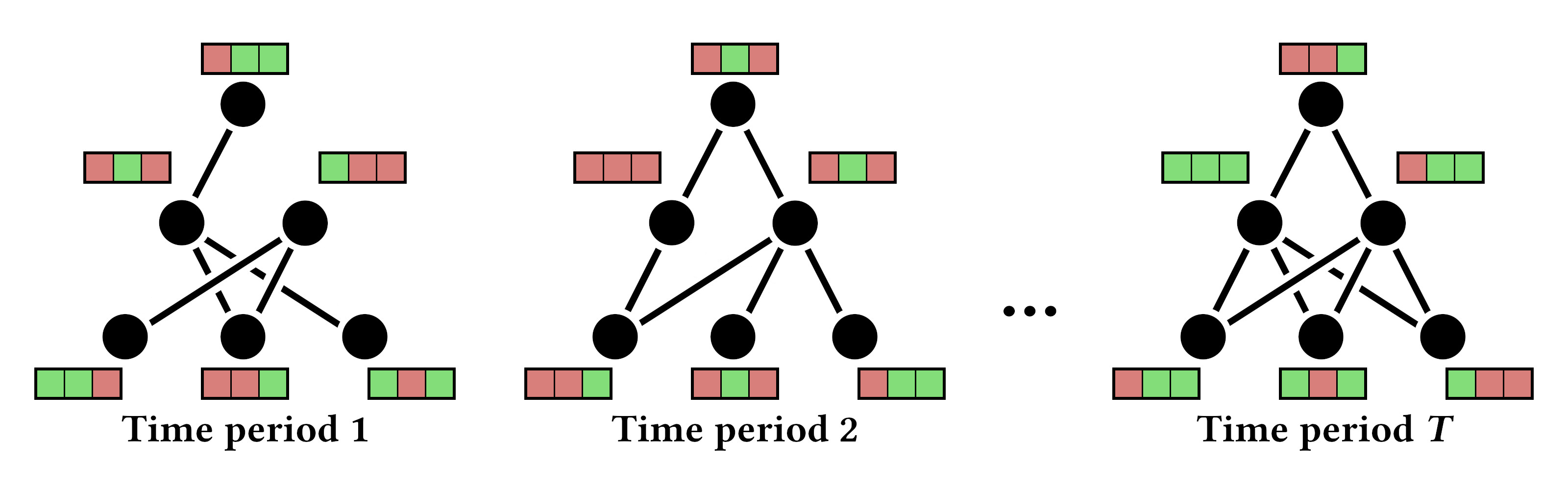}}

\subfloat[Dynamic graph with static signal.]{\includegraphics[height=1.0in]{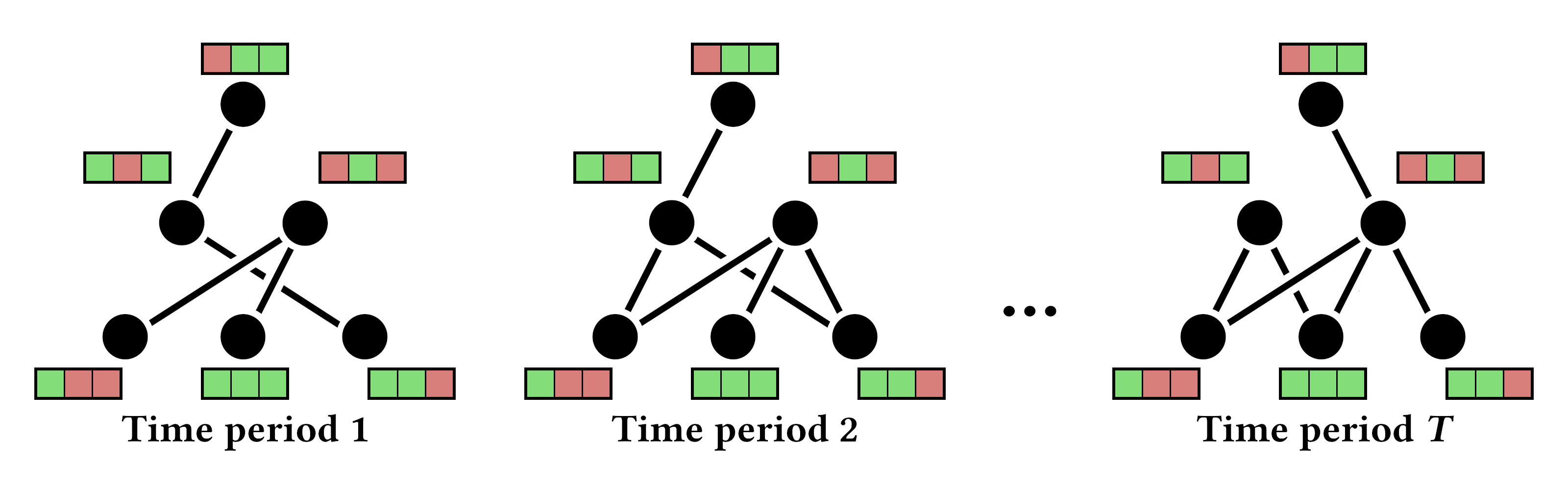}}

\subfloat[Static graph with temporal signal.]{\includegraphics[height=1.0in]{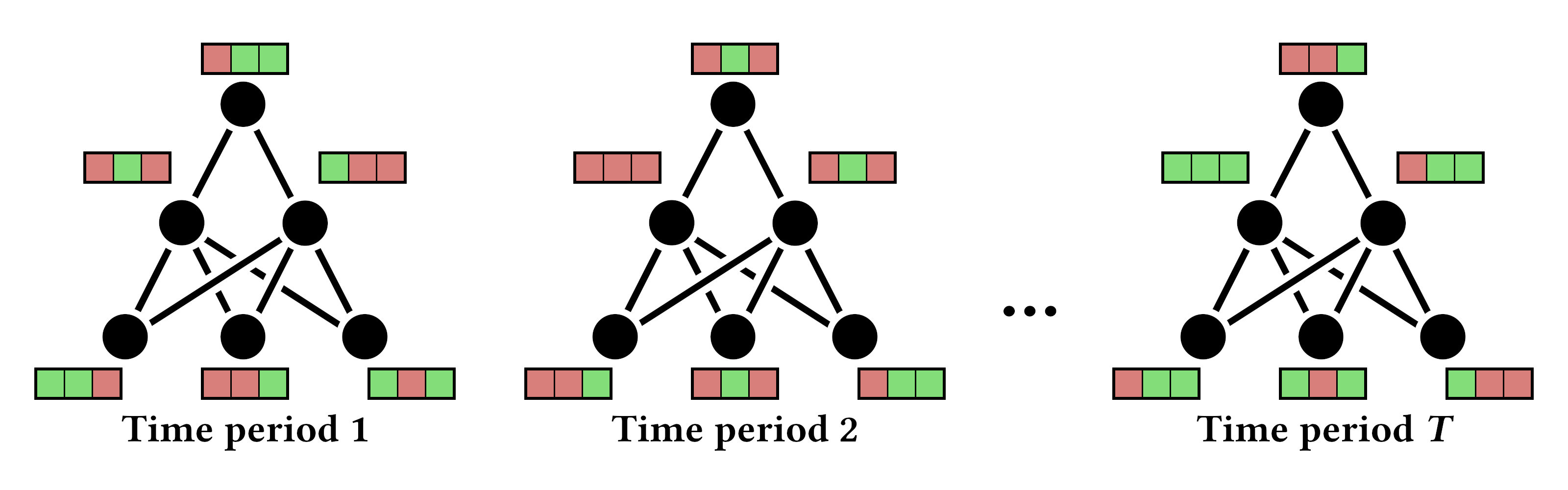}}
\caption{The data iterators in PyTorch Geometric Temporal can provide temporal snapshots for all of the non static geometric deep learning scenarios.}\label{fig:dynamics}
\end{figure}

Our framework considers specific input data types on which the spatiotemporal machine learning models operate. Input data types can differ in terms of the dynamics of the graph and that of the modelled vertex attributes. We take a discrete temporal snapshot view of this data representation problem  \cite{holme2012temporal, holme2015modern} and our work considers three spatiotemporal data types which can be described by the subplots of Figure \ref{fig:dynamics} and the following formal definitions:
\begin{definition} \textbf{Dynamic graph with temporal signal} A dynamic graph with a temporal signal is the ordered set of graph and node feature matrix tuples  $\mathcal{D}=\left\{ ( \mathcal{G}_1, \mathbf{X}_1),\dots, (\mathcal{G}_T, \mathbf{X}_T)\right \}$ where  the vertex sets satisfy that $V_t=V,\,\,\forall t \in \left\{1,\dots,T\right\}$ and  the node feature matrices that $\mathbf{X}_t\in \mathbb{R}^{|V|\times d},\,\,\forall t \in \left\{1,\dots,T\right\}.$

\end{definition} 

\begin{definition} \textbf{Dynamic graph with static signal.} A dynamic graph with a static signal is the ordered set of graph and node feature matrix tuples  $\mathcal{D}=\left\{ ( \mathcal{G}_1, \mathbf{X}),\dots, (\mathcal{G}_T, \mathbf{X})\right \}$ where vertex sets satisfy $V_t=V,$ $\forall t \in \left\{1,\dots,T\right\}$ and the node feature matrix that $\mathbf{X}\in \mathbb{R}^{|V|\times d}.$

\end{definition}
\begin{definition}   \textbf{Static graph with temporal signal.}  A static graph with a temporal signal is the ordered set of graph and node feature matrix tuples  $\mathcal{D}=\left\{ ( \mathcal{G}, \mathbf{X}_1),\dots, (\mathcal{G}, \mathbf{X}_T)\right \}$ where  the node feature matrix satisfies that $\mathbf{X}_t\in \mathbb{R}^{|V|\times d},\,\,\forall t \in \left\{1,\dots,T\right\}.$
\end{definition}

Representing spatiotemporal data based on these theoretical concepts allows us the creation of memory efficient data structures which conceptualize these definitions in practice well.

\subsection{Deep Learning with Time and Geometry}
Our work provides deep learning models that operate on data which has both temporal and spatial aspects. These techniques are natural recombinations of existing neural network layers that operate on sequences and static graph-structured data.
\subsubsection{Temporal Deep Learning} A large family of temporal deep learning models such as the \textit{LSTM}  \cite{lstm} and \textit{GRU} \cite{gru} generates in-memory representations of data points which are iteratively updated as it learns by new snapshots. Another family of deep learning models uses the \textit{attention mechanism} \cite{luong2015effective,bahdanau2015neural,vaswani2017attention} to learn representations of the data points which are adaptively recontextualized based on the temporal history. These types of models serve as templates for the temporal block of spatiotemporal deep learning models.
\subsubsection{Static Graph Representation Learning} Learning representations of vertices, edges and whole graphs with graph neural networks in a supervised or unsupervised way can be described by the \textit{message passing} formalism \cite{gilmer2017neural}. In this conceptual framework using the node and edge attributes in a graph as parametric function generates compressed representations (messages) which are propagated between the nodes based on a message-passing rule and aggregated to form new representations.  Most of the existing graph neural network architectures such as GCN \cite{kipf2017semi}, GGCN\cite{gated}, ChebyConv \cite{chebyshevconv}, and RGCN \cite{schlichtkrull2018modeling} fit perfectly into this general description of graph neural networks. Models are differentiated by assumptions about the input graph (e.g. node heterogeneity, multiplexity, presence of edge attributes), the message compression function used,  the propagation scheme and message aggregation function applied to the received messages.

\begin{table}[h!]
\centering
\caption{A comparison of spatiotemporal deep learning models in PyTorch Geometric Temporal based on the temporal and spatial block, order of proximity considered and the hetereogeneity of the edges.}\label{tab:methods}
{
\small
\begin{tabular}{lcccc}

    \textbf{Model}    & \specialcell{\textbf{Temporal}\\\textbf{Layer}}   & \specialcell{\textbf{GNN}\\ \textbf{Layer}}    & \specialcell{\textbf{Proximity}\\ \textbf{Order}}    & \specialcell{\textbf{Multi}\\ \textbf{Type}}   \\ \hline
\textbf{DCRNN} \cite{li2018diffusion}&   GRU   &   DiffConv    &  Higher    &   False       \\[0.1cm]
\textbf{GConvGRU} \cite{gconvlstm}&GRU   &   Chebyshev    &  Lower    &   False      \\[0.1cm]
\textbf{GConvLSTM} \cite{gconvlstm}&LSTM     &   Chebyshev    &  Lower    &   False  \\[0.1cm]
\textbf{GC-LSTM} \cite{gclstm}&   LSTM   & Chebyshev      &  Lower    &   True \\[0.1cm]
\textbf{DyGrAE} \cite{dyngrae_1,dyggnn}&  LSTM &   GGCN    &  Higher    &   False    \\[0.1cm]
\textbf{LRGCN} \cite{li2019predicting}&LSTM     &  RGCN     &  Lower    &   False       \\[0.1cm]
\textbf{EGCN-H} \cite{evolvegcn}&GRU     &   GCN    &  Lower    &   False        \\[0.1cm]

\textbf{EGCN-O} \cite{evolvegcn}&   LSTM   &   GCN    &  Lower    &   False       \\[0.1cm]
\textbf{T-GCN} \cite{tgcn}&GRU   &   GCN    &  Lower     &   False  \\[0.1cm]
\textbf{A3T-GCN} \cite{a3tgcn}&GRU   &   GCN    &  Lower     &   False  \\[0.1cm]
\textbf{AGCRN} \cite{bai2020adaptive} &  GRU &  Chebyshev   & Higher    &False    \\[0.1cm]
\textbf{MPNN LSTM} \cite{panagopoulos2020transfer} &  LSTM  &  GCN   & Lower    &False    \\[0.1cm]
\textbf{STGCN} \cite{yu2018spatio}&  Attention   &   Chebyshev    &  Higher    &   False     \\[0.1cm]
\textbf{ASTGCN} \cite{guo2019attention}&  Attention   &   Chebyshev    &  Higher    &   False     \\[0.1cm]
\textbf{MSTGCN} \cite{guo2019attention}&  Attention   &   Chebyshev    &  Higher    &   False     \\[0.1cm]
\textbf{GMAN} \cite{zheng2020gman}&  Attention   &  Custom    &  Lower    &   False    \\[0.1cm]
\textbf{MTGNN} \cite{wu2020connecting} &  Attention  &  Custom   & Higher    &False    \\[0.1cm]
\textbf{AAGCN} \cite{shi2019two} &  Attention  &  Custom   & Higher    &False    \\[0.1cm]
\hline
\end{tabular}
}
\end{table}

\subsubsection{Spatiotemporal Deep Learning} A spatiotemporal deep learning model fuses the basic conceptual ideas of temporal deep learning techniques and graph representation learning. Operating on a temporal graph sequence these models perform message-passing at each time point with a graph neural network block and the new temporal information is incorporated by a temporal deep learning block. This design allows for sharing salient temporal and spatial autocorrelation information across the spatial units.  The temporal and spatial layers which are fused together in a single parametric machine learning model are trained together jointly by exploiting the fact that the fused models are end-to-end differentiable. In Table \ref{tab:methods} we summarized the spatiotemporal deep learning models implemented in the framework which we categorized based on the temporal and graph neural network layer blocks, the order of spatial proximity and heterogeneity of the edge set.

\subsection{Graph Representation Learning Software}
The current graph representation learning software ecosystem which allows academic research and industrial deployment extends open-source auto-differentiation libraries such as TensorFlow \cite{abadi2016tensorflow}, PyTorch \cite{paszke2019pytorch}, MxNet \cite{mxnet} and JAX \cite{jax,frostig2018compiling}. Our work does the same as we build on the PyTorch Geometric ecosystem. We summarized the characteristics of these libraries in Table \ref{tab:lib_comparison} which allows for comparing frameworks based on the backend, presence of supervised training functionalities, presence of temporal models and GPU support. Our proposed framework is the only one to date which allows the supervised training of temporal graph representation learning models with graphics card based acceleration.

\begin{table}[h!]
\centering
\caption{A desiderata and automatic differentiation backend library based comparison of open-source geometric deep learning libraries.}\label{tab:lib_comparison}
{\small\begin{tabular}{lccccc}
\hline
\textbf{Library} & \textbf{Backend}           & \textbf{Supervised} & \textbf{Temporal} & \textbf{GPU} \\
\hline
PT Geometric   \cite{pytorch_geometric} &PT&      \cmark       &      \xmark         &  \cmark \\
Geometric2DR   \cite{geometric2dr} &PT&      \xmark       &      \xmark         &  \cmark \\
CogDL   \cite{cen2021cogdl} &PT&      \cmark       &      \xmark         &  \cmark \\
Spektral   \cite{spektral} &TF&      \cmark       &      \xmark         &  \cmark \\
TF Geometric  \cite{hu2021efficient} &TF&      \cmark       &      \xmark         &  \cmark \\
StellarGraph   \cite{StellarGraph} &TF&      \cmark       &      \xmark         &  \cmark \\
DGL   \cite{dlg} &TF/PT/MX&      \cmark       &      \xmark         &  \cmark \\
DIG   \cite{liu2021dig} &PT&      \cmark       &      \xmark         &  \cmark \\
Jraph  \cite{jraph2020github}& JAX&      \cmark       &      \xmark         &  \cmark \\
Graph-Learn \cite{euler}  &Custom &      \cmark       &      \xmark         &  \cmark \\
GEM   \cite{goyal2018dynamicgem} &TF&      \xmark       &      \xmark         &  \cmark \\
DynamicGEM   \cite{goyal2018gem} &TF&      \xmark       &      \cmark         &  \cmark \\
OpenNE  \cite{openne} &Custom&      \xmark       &      \xmark         &  \xmark \\
Karate Club   \cite{rozemberczki2020karate} &Custom&      \xmark       &      \xmark         &  \xmark \\
\hline
Our Work &PT&      \cmark       &      \cmark         &  \cmark \\
\hline
\end{tabular}}
\end{table}

\subsection{Spatiotemporal Data Analytics Software}
The open-source ecosystem for spatiotemporal data processing consists of specialized database systems, basic analytical tools and advanced machine learning libraries. We summarized the characteristics of the most popular libraries in Table \ref{tab:spat_lib_comparison} with respect to the year of release, purpose of the framework, source code language and GPU support.

First, it is evident that most spatiotemporal data processing tools are fairly new and there is much space for contributions in each subcategory. Second, the database systems are written in high-performance languages while the analytics and machine learning oriented tools have a pure Python/R design or a wrapper written in these languages. Finally, the use of GPU acceleration is not widespread which alludes to the fact that current spatiotemporal data processing tools might have a scalability issue. Our proposed framework PyTorch Geometric Temporal is the first fully open-source GPU accelerated spatiotemporal machine learning library. 

\begin{table}[h!]
\centering
\caption{A multi-aspect comparison of open-source spatiotemporal database systems, data analytics libraries and machine learning frameworks.}\label{tab:spat_lib_comparison}
{\small\begin{tabular}{lccccc}
\hline
\textbf{Library} & \textbf{Year}           & \textbf{Purpose} & \textbf{Language} & \textbf{GPU} \\
\hline
GeoWave \cite{whitby2017geowave} &2016&      Database     &      Java       &  \xmark \\
StacSpec \cite{hanson2019open} &2017&     Database      &     Javascript        &  \xmark \\
MobilityDB \cite{zimanyi2020mobilitydb} &2019&     Database      &      C        &  \xmark \\
PyStac \cite{pystac} &2020&      Database    &    Python       &  \xmark \\
StaRs \cite{stars} &2017&      Analytics      &      R         &  \xmark \\
CuSpatial \cite{cuspatial} &2019&      Analytics      &      Python         &  \cmark \\
PySAL \cite{rey2010pysal} &2017&      Machine Learning       &      Python         &  \xmark \\
STDMTMB \cite{sdmTMB} &2018&      Machine Learning     &      R      &  \xmark \\
\hline 
Our work &2021&     Machine Learning      &      Python       &  \cmark \\

\hline
\end{tabular}}
\end{table}

\section{The Framework design}\label{sec:design}
Our primary goal is to give a general theoretical overview of the framework, discuss the framework design choices, give a detailed practical example and highlight our strategy for the long term viability  and maintenance of the project.
\subsection{Neural Network Layer Design}
The spatiotemporal neural network layers are implemented as classes in the framework. Each of the classes has a similar architecture driven by a few simple design principles.
\subsubsection{Non-proliferation of classes} The framework reuses the existing high level neural network layer classes as building blocks from the PyTorch and PyTorch Geometric ecosystems. The goal of the library is not to replace the existing frameworks. This design strategy makes sure that the number of auxiliary classes in the framework is kept low and that the framework interfaces well with the rest of the ecosystem. 
\subsubsection{Hyperparameter inspection and type hinting} The neural network layers do not have default hyperparameter settings as some of these have to be set in a dataset dependent manner. In order to help with this, the layer hyperparameters are stored as public class attributes and they are available for inspection. Moreover, the constructors of the neural network layers use type hinting which helps the end-users to set the hyperparameters.

\subsubsection{Limited number of public methods} The spatiotemporal neural network layers in our framework have a limited number of public methods for simplicity. For example, the auxiliary layer initialization methods and other internal model mechanics are implemented as private methods. All of the layers provide a forward method and those which explicitly use the \textit{message-passing} scheme in PyTorch Geometric provide a public \textit{message} method.

\subsubsection{Auxiliary layers} The auxiliary neural network layers which are not part of the \textit{PyTorch Geometric} ecosystem such as diffusion convolutional graph neural networks \cite{li2018diffusion} are implemented as standalone neural network layers in the framework. These layers are available for the design of novel neural network architectures as individual components.

\subsection{Data Structures}
The design of \textit{PyTorch Geometric Temporal} required the introduction of custom data structures which can efficiently store the datasets and provide temporally ordered snapshots for batching.
\subsubsection{Spatiotemporal Signal Iterators} Based on the categorization of spatiotemporal signals discussed in Section \ref{sec:related_work} we implemented three types of \textit{Spatiotemporal Signal Iterators}. These iterators store spatiotemporal datasets in memory efficiently without redundancy. For example a \textit{Static Graph Temporal Signal} iterator will not store the edge indices and weights for each time period in order to save memory. By iterating over a \textit{Spatiotemporal Signal Iterator} at each step a graph snapshot is returned which describes the graph of interest at a given point in time. Graph snapshots are returned in temporal order by the iterators. The \textit{Spatiotemporal Signal Iterators} can be indexed directly to access a specific graph snapshot -- a design choice which allows the use of advanced temporal batching. 

\subsubsection{Graph Snapshots} The time period specific snapshots which consist of labels, features, edge indices and weights are stored as \textit{NumPy} arrays \cite{van2011numpy} in memory, but returned as a \textit{PyTorch Geometric Data} object instance \cite{pytorch_geometric} by the \textit{Spatiotemporal Signal Iterators} when these are iterated on. This design choice hedges against the proliferation of classes and exploits the existing and widely used compact data structures from the \textit{PyTorch} ecosystem \cite{pytorch}. 
\subsubsection{Train-Test Splitting} As part of the library we provide a temporal train-test splitting function which creates train and test snapshot iterators from a Spatiotemporal Signal Iterator given a test dataset ratio. This parameter of the splitting function decides the fraction of data that is separated from the end of the spatiotemporal graph snapshot sequence for testing. The returned iterators have the same type as the input iterator. Importantly, this splitting does not influence the applicability of widely used semi-supervised model training strategies such as node masking. 
\subsubsection{Integrated Benchmark Dataset Loaders} We provided easy-to-use practical data loader classes for widely used existing \cite{,panagopoulos2020transfer} and the newly released benchmark datasets. These loaders return \textit{Spatiotemporal Signal Iterators} which can be used for training existing and custom designed spatiotemporal neural network architectures to solve supervised machine learning problems. 
\subsection{Design in Practice Case Study: Cumulative Model Training on CPU}\label{subsec:cumulative}
In the following we overview a simple end-to-end machine learning pipeline designed with PyTorch Geometric Temporal. These code snippets solve a practical epidemiological forecasting problem -- predicting the weekly number of chickenpox cases in Hungary \cite{chickenpox}. The pipeline consists of data preparation, model definition, training and evaluation phases.
\begin{code}
\begin{minted}[linenos,fontsize=\small,xleftmargin=0.5cm,numbersep=3pt,frame=lines]{python}
from torch_geometric_temporal import ChickenpoxDatasetLoader
from torch_geometric_temporal import temporal_signal_split

loader = ChickenpoxDatasetLoader()

dataset = loader.get_dataset()

train, test = temporal_signal_split(dataset,
                                    train_ratio=0.9)
\end{minted}
\captionof{listing}{Loading a default benchmark dataset and creating a temporal split with PyTorch Geometric Temporal}\label{code:split_example}
\end{code}
\subsubsection{Dataset Loading and Splitting} In Listings \ref{code:split_example} as a first step we import the Hungarian chickenpox cases benchmark dataset loader and the temporal train test splitter function (lines 1-2). We define the dataset loader (line 4) and use the \textit{get\_dataset()} class method to return a temporal signal iterator (line 5). Finally, we create a train-test split of the spatiotemporal dataset by using the splitting function and retain 10\% of the temporal snapshots for model performance evaluation (lines 7-8).

\begin{code}

\begin{minted}[linenos,fontsize=\small,xleftmargin=0.5cm,numbersep=3pt,frame=lines]{python}
import torch
import torch.nn.functional as F
from torch_geometric_temporal.nn.recurrent import DCRNN

class RecurrentGCN(torch.nn.Module):
    def __init__(self, node_features, filters):
        super(RecurrentGCN, self).__init__()
        self.recurrent = DCRNN(node_features, filters, 1)
        self.linear = torch.nn.Linear(filters, 1)

    def forward(self, x, edge_index, edge_weight):
        h = self.recurrent(x, edge_index, edge_weight)
        h = F.relu(h)
        h = F.dropout(h, training=self.training)
        h = self.linear(h)
        return h
\end{minted}
\captionof{listing}{Defining a recurrent graph convolutonal neural network using PyTorch Geometric Temporal consisting of a diffusion convolutional spatiotemporal layer followed by rectified linear unit activations, dropout and a feedforward neural network layer.}\label{code:model_definition}
\end{code}

\subsubsection{Recurrent Graph Convolutional Model Definition} We define a recurrent graph convolutional  neural network model in Listings \ref{code:model_definition}. We import the base and functional programming PyTorch libraries and one of the neural network layers from PyTorch Geometric Temporal  (lines 1-3). The model requires a node feature count and convolutional filter parameter in the constructor (line 6). The model consists of a one-hop Diffusion Convolutional Recurrent Neural Network layer \cite{li2018diffusion} and a fully connected layer with a single neuron (lines 8-9). 

In the forward pass method of the neural network the model uses the vertex features, edges and the optional edge weights (line 11). The initial recurrent graph convolution based aggregation (line 12) is followed by a rectified linear unit activation function \cite{nair2010rectified} and dropout \cite{srivastava2014dropout} for regularization (lines 13-14). Using the fully-connected layer the model outputs a single score for each spatial unit (lines 15-16).
\begin{code}
\begin{minted}[linenos,fontsize=\small,xleftmargin=0.5cm,numbersep=3pt,frame=lines]{python}
model = RecurrentGCN(node_features=8, filters=32)

optimizer = torch.optim.Adam(model.parameters(), lr=0.01)

model.train()

for epoch in range(200):
    cost = 0
    for time, snapshot in enumerate(train):
        y_hat = model(snapshot.x,
                      snapshot.edge_index,
                      snapshot.edge_attr)
        cost = cost + torch.mean((y_hat-snapshot.y)**2)
    cost = cost / (time+1)
    cost.backward()
    optimizer.step()
    optimizer.zero_grad()
\end{minted}
\captionof{listing}{Creating a recurrent graph convolutional neural network instance and training it by cumulative weight updates.}\label{code:training}
\end{code}
\subsubsection{Model Training} Using the dataset split and the model definition we can turn our attention to training a regressor. In Listings \ref{code:training} we create a model instance (line 1), transfer the model parameters (line 3) to the Adam optimizer \cite{kingma_adam_2014} which uses a learning rate of 0.01 and set the model to be trainable (line 5). In each epoch we set the accumulated cost to be zero (line 8) iterate over the temporal snapshots in the training data (line 9), make forward passes with the model on each temporal snapshot and accumulate the spatial unit specific mean squared errors (lines 10-13). We normalize the cost, backpropagate and update the model parameters (lines 14-17).

\begin{code}
\begin{minted}[linenos,fontsize=\small,xleftmargin=0.5cm,numbersep=3pt,frame=lines]{python}
model.eval()
cost = 0
for time, snapshot in enumerate(test):
    y_hat = model(snapshot.x,
                  snapshot.edge_index,
                  snapshot.edge_attr)
    cost = cost + torch.mean((y_hat-snapshot.y)**2)
cost = cost / (time+1)
cost = cost.item()
print("MSE: {:.4f}".format(cost))
\end{minted}
\captionof{listing}{Evaluating the recurrent graph convolutional neural network on the test portion of the spatiotemporal dataset using the time unit averaged mean squared error.}\label{code:evaluation}
\end{code}

\subsubsection{Model Evaluation} The scoring of the trained recurrent graph neural network in Listings \ref{code:evaluation} uses the snapshots in the test dataset. We set the model to be non trainable and the accumulated squared error as zero (lines 1-2). We iterate over the test spatiotemporal snapshots, make forward passes to predict the number of chickenpox cases and accumulate the squared error (lines 3-7). The accumulated errors are normalized and we can print the mean squared error  calculated on the whole test horizon (lines 8-10).

\subsection{Design in Practice Case Study: Incremental Model Training with GPU Acceleration}
Exploiting the power of GPU based acceleration of computations happens at the training and evaluation steps of the PyTorch Geometric Temporal pipelines. In this case study we assume that the Hungarian Chickenpox cases dataset is already loaded in memory, the temporal split happened and a model class was defined by the code snippets in Listings \ref{code:split_example} and \ref{code:model_definition}. Moreover, we assume that the machine used for training the neural network can access a single CUDA compatible GPU device \cite{sanders2010cuda}.

\begin{code}
\begin{minted}[linenos,fontsize=\small,xleftmargin=0.5cm,numbersep=3pt,frame=lines]{python}
model = RecurrentGCN(node_features=8, filters=32)
device = torch.device('cuda')
model = model.to(device)

optimizer = torch.optim.Adam(model.parameters(), lr=0.01)
model.train()

for epoch in range(200):
    for snapshot in train:
        snapshot = snapshot.to(device)
        y_hat = model(snapshot.x,
                      snapshot.edge_index,
                      snapshot.edge_attr)
        cost = torch.mean((y_hat-snapshot.y)**2)
        cost.backward()
        optimizer.step()
        optimizer.zero_grad()
\end{minted}
\captionof{listing}{Creating a recurrent graph convolutional neural network instance and training it by incremental weight updates on a GPU.}\label{code:training_2}
\end{code}

\subsubsection{Model Training} 
In Listings \ref{code:training_2} we demonstrate accelerated training with incremental weight updates. 
The model of interest and the device used for training are defined while the model is transferred to the GPU (lines 1-3). The optimizer registers the model parameters and the model parameters are set to be trainable (lines 5-6). We iterate over the temporal snapshot iterator 200 times and the iterator returns a temporal snapshot in each step. Importantly the snapshots which are PyTorch Geometric Data objects are transferred to the GPU (lines 8-10). The use of PyTorch Geometric Data objects as temporal snapshots allows the transfer of the time period specific edges, node features and target vector with a single command. Using the input data a forward pass is made, loss is accumulated and weight updates happen using the optimizer in each time period (lines 11-17). Compared to the cumulative backpropagation based training approach discussed in Subsection \ref{subsec:cumulative} this backpropagation strategy is slower as weight updates happen at each time step, not just at the end of training epochs.

\subsubsection{Model Evaluation} During model scoring the GPU can be utilized again. The snippet in Listings \ref{code:evaluation_2} demonstrates that the only modification needed for accelerated evaluation is the transfer of snapshots to the GPU. In each time period we move the temporal snapshot to the device to do the forward pass (line 4). We do the forward pass with the model and the snapshot on the GPU and accumulate the loss (lines 5-8). The loss value is averaged out and detached from the GPU for printing (lines 9-11).

\begin{code}
\begin{minted}[linenos,fontsize=\small,xleftmargin=0.5cm,numbersep=3pt,frame=lines]{python}
model.eval()
cost = 0
for time, snapshot in enumerate(test):
    snapshot = snapshot.to(device)
    y_hat = model(snapshot.x,
                  snapshot.edge_index,
                  snapshot.edge_attr)
    cost = cost + torch.mean((y_hat-snapshot.y)**2)
cost = cost / (time+1)
cost = cost.item()
print("MSE: {:.4f}".format(cost))
\end{minted}
\captionof{listing}{Evaluating the recurrent graph convolutional neural network with GPU based acceleration.}\label{code:evaluation_2}
\end{code}
\subsection{Maintaining PyTorch Geometric Temporal}
The viability of the project is made possible by the open-source code, version control, public releases, automatically generated documentation, continuous integration, and near 100\% test coverage.
\subsubsection{Open-Source Code-Base and Public Releases} The source code of \textit{PyTorch Geometric Temporal} is publicly available on \textit{GitHub} under the MIT license. Using an open version control system allowed us to have a large group collaborate on the project and have external contributors who also submitted feature requests. The public releases of the library are also made available on the \textit{Python Package Index}, which means that the framework can be installed via the \textit{pip} command using the terminal.
\subsubsection{Documentation} The source-code of \textit{PyTorch Geometric Temporal} and \textit{Sphinx} \cite{brandl2010sphinx} are used to generate a publicly available documentation of the library at \url{https://pytorch-geometric-temporal.readthedocs.io/}. This documentation is automatically created every time when the code-base changes in the public repository. The documentation covers the constructors and public methods of neural network layers, temporal signal iterators, public dataset loaders and splitters. It also includes a list of relevant research papers, an in-depth installation guide, a detailed getting-started tutorial and a list of integrated benchmark datasets.

\subsubsection{Continuous Integration} We provide continuous integration for \textit{PyTorch Geometric Temporal} with \textit{GitHub Actions} which are available for free on \textit{GitHub} without limitations on the number of builds. When the code is updated on any branch of the repository the build process is triggered and the library is deployed on \textit{Linux}, \textit{Windows} and \textit{macOS} virtual machines. 
\subsubsection{Unit Tests and Code Coverage}
The temporal graph neural network layers, custom data structures and benchmark dataset loaders are all covered by unit tests. These unit tests can be executed locally using the source code. Unit tests are also triggered by the continuous integration provided by \textit{GitHub Actions}. When the master branch of the open-source \textit{GitHub} repository is updated, the build is successful, and all of the unit tests pass a coverage report is generated by \textit{CodeCov}.

\section{Experimental evaluation}\label{sec:experiments}
The proposed framework is evaluated on node level regression tasks using novel datasets which we release with the paper. We also evaluate the effect of various batching techniques on the predictive performance and runtime.

\subsection{Datasets}
We release new spatiotemporal benchmark datasets with \textit{PyTorch Geometric Temporal} which can be used to test models on node level regression tasks. The descriptive statistics and properties of these newly introduced benchmark datasets are summarized in Table \ref{tab:desc_discrete}.

\begin{table}[h!]
\centering
\small
\caption{Properties and granularity of the spatiotemporal datasets introduced in the paper with information about the number of time periods ($T$) and spatial units ($|V|$).}\label{tab:desc_discrete}
{
\setlength{\tabcolsep}{2pt}

\begin{tabular}{cccccc}
\hline
\textbf{Dataset} & \textbf{Signal} & \textbf{Graph} & \textbf{Frequency} & $T$ & $|V|$ \\
\hline
    \specialcell{Chickenpox Hungary} & Temporal&Static & Weekly & 522 & 20 \\
     \specialcell{Windmill Large} & Temporal & Static & Hourly & 17,472 & 319 \\
     \specialcell{Windmill Medium} & Temporal & Static & Hourly &  17,472  &26 \\
     \specialcell{Windmill Small} & Temporal & Static & Hourly & 17,472  &11 \\
     \specialcell{Pedal Me Deliveries} & Temporal & Static & Weekly & 36 & 15 \\
 \specialcell{Wikipedia Math} & Temporal&Static&Daily & 731 & 1,068 \\
 \specialcell{Twitter Tennis RG} & Static & Dynamic & Hourly & 120 & 1000 \\
  \specialcell{Twitter Tennis UO} & Static & Dynamic & Hourly & 112 & 1000 \\
  \specialcell{Covid19 England} & Temporal & Dynamic & Daily & 61 & 129 \\
  \specialcell{Montevideo Buses} & Temporal & Static & Hourly & 744 & 675 \\
  \specialcell{MTM-1 Hand Motions} & Temporal & Static & 1/24 Seconds & 14,469 & 21 \\
 \hline
\end{tabular}
}
\end{table}

These newly released datasets are the following:
\begin{itemize}
    \item \textbf{Chickenpox Hungary.} A spatiotemporal dataset about the officially reported cases of chickenpox in Hungary. The nodes are counties and edges describe direct neighbourhood relationships. The dataset covers the weeks between 2005 and 2015 without missingness. 
    \item \textbf{Windmill Output Datasets.} An hourly windfarm energy output dataset covering 2 years from a European country. Edge weights are calculated from the proximity of the windmills -- high weights imply that two windmill stations are in close vicinity. The size of the dataset relates to the groupping of windfarms considered; the smaller datasets are more localized to a single region.
    \item \textbf{Pedal Me Deliveries.} A dataset about the number of weekly bicycle package deliveries by Pedal Me in London during 2020 and 2021. Nodes in the graph represent geographical units and edges are proximity based mutual adjacency relationships.
    \item \textbf{Wikipedia Math.} Contains Wikipedia pages about popular mathematics topics and edges describe the links from one page to another. Features describe the number of daily visits between 2019 and 2021 March.
    \item \textbf{Twitter Tennis RG and UO.} Twitter mention graphs of major tennis tournaments from 2017. Each snapshot contains the graph of popular player or sport news accounts and mentions between them \cite{Beres2018,Beres2019}. Node labels encode the number of mentions received and vertex features are structural properties.
    \item \textbf{Covid19 England.} A dataset about mass mobility between regions in England and the number of confirmed COVID-19 cases from March to May 2020 \cite{panagopoulos2020transfer}. Each day contains a different mobility graph and node features corresponding to the number of cases in the previous days. Mobility stems from Facebook Data For Good \footnote{\url{ https://dataforgood.fb.com/}} and cases from gov.uk \footnote{\url{https://coronavirus.data.gov.uk/}}.
    \item \textbf{Montevideo Buses.} A dataset about the hourly passenger inflow at bus stop level for eleven bus lines from the city of Montevideo. Nodes are bus stops and edges represent connections between the stops; the dataset covers a whole month of traffic patterns.
    \item \textbf{MTM-1 Hand Motions.} A temporal dataset of Methods-Time Measurement-1 \cite{Maynard1948} motions, signalled as consecutive graph frames of 21 3D hand key points that were acquired via MediaPipe Hands \cite{zhang2020mediapipe} from original RGB-Video material. Node features encode the normalized xyz-coordinates of each finger joint and the vertices are connected according to the human hand structure.
\end{itemize}

\begin{table*}[h!]
\centering
\caption{The predictive performance of spatiotemporal neural networks evaluated by average mean squared error. We report average performances calculated from 10 experimental repetitions with standard deviations around the average mean squared error calculated on 10\% forecasting horizons. We use the incremental and cumulative backpropagation strategies.}\label{tab:predictive_performance}
{
\small
\begin{tabular}{lcccccccc}

        & \multicolumn{2}{c}{\textbf{Chickenpox Hungary}}& \multicolumn{2}{c}{\textbf{Twitter Tennis RG}} & \multicolumn{2}{c}{\textbf{PedalMe London}} & \multicolumn{2}{c}{\textbf{Wikipedia Math}} \\
\cmidrule[0.4pt](lr{0.125em}){2-3}
\cmidrule[0.4pt](lr{0.125em}){4-5}
\cmidrule[0.4pt](lr{0.125em}){6-7}
\cmidrule[0.4pt](lr{0.125em}){8-9}
       & Incremental     & Cumulative     & Incremental     & Cumulative    & Incremental     & Cumulative     & Incremental     & Cumulative \\ \hline
\textbf{DCRNN} \cite{li2018diffusion}&    $1.124\pm 0.015$      &    $1.123\pm 0.014$       &   $2.049\pm 0.023$       &    $2.043\pm 0.016$       &      $1.463\pm 0.019$     &        $1.450\pm 0.024$  &     $\mathbf{0.679\pm 0.020}$      &     $\mathbf{0.803\pm 0.018}$        \\[0.1cm]
\textbf{GConvGRU} \cite{gconvlstm}& $1.128\pm 0.011$      &    $1.132\pm 0.023$       &   $2.051\pm 0.020$       &    $2.007\pm 0.022$       &      $1.622\pm 0.032$     &        $1.944\pm 0.013$  &     $0.657\pm 0.015$      &     $0.837\pm 0.021$        \\[0.1cm]
\textbf{GConvLSTM} \cite{gconvlstm}&$1.121\pm 0.014$      &    $1.119\pm 0.022$       &   $2.049\pm 0.024$       &    $2.007\pm 0.012$       &      $1.442\pm 0.028$     &        $1.433\pm 0.020$  &     $0.777\pm 0.021$      &     $0.868\pm 0.018$        \\[0.1cm]
\textbf{GC-LSTM} \cite{gclstm}&    $1.115\pm 0.014$      &    $1.116\pm 0.023$       &   $2.053\pm 0.024$       &    $2.032\pm 0.015$       &      $\mathbf{1.455\pm 0.023}$     &        $1.468\pm 0.025$  &     $0.779\pm 0.023$      &     $0.852\pm 0.016$        \\[0.1cm]
\textbf{DyGrAE} \cite{dyngrae_1,dyggnn}& $1.120\pm 0.021$      &    $1.118\pm 0.015$       &   $\mathbf{2.031\pm 0.006}$       &    $2.007\pm 0.004$       &      $\mathbf{1.455\pm 0.031}$     &        $1.456\pm 0.019$  &     $0.773\pm 0.009$      &     $0.816\pm 0.016$        \\[0.1cm]
\textbf{EGCN-H} \cite{evolvegcn}&$\mathbf{1.113\pm 0.016}$      &    $\mathbf{1.104\pm 0.024}$       &   $2.040\pm 0.018$       &    $\mathbf{2.006\pm 0.008}$       &      $1.467\pm 0.026$     &        $1.436\pm 0.017$  &     $0.775\pm 0.022$      &     $0.857\pm 0.022$        \\[0.1cm]

\textbf{EGCN-O} \cite{evolvegcn}&    $1.124\pm 0.009$      &    $1.119\pm 0.020$       &   $2.055\pm 0.020$       &    $2.010\pm 0.014$       &      $1.491\pm 0.024$     &        $\mathbf{1.430\pm 0.023}$  &     $0.750\pm 0.014$      &     $0.823\pm 0.014$        \\[0.1cm]
\textbf{A3T-GCN}\cite{a3tgcn}&$1.114\pm 0.008$      &    $1.119\pm 0.018$       &   $2.045\pm 0.021$       &    $2.008\pm 0.016$       &      $1.469\pm 0.027$     &        $1.475\pm 0.029$  &     $0.781\pm 0.011$      &     $0.872\pm 0.017$        \\[0.1cm]
\textbf{T-GCN} \cite{tgcn}& $1.117\pm 0.011$      &  $1.111\pm 0.022$  &   $2.045\pm 0.027$       &   $2.008\pm 0.017$            &      $1.479\pm 0.012$     &        $1.481\pm 0.029$  &     $0.764\pm 0.011$      &     $0.846\pm 0.020$        \\[0.1cm]
\textbf{MPNN LSTM} \cite{panagopoulos2020transfer}&$1.116\pm 0.023$      &    $1.129\pm 0.021$       &   $2.053\pm 0.041$       &    $2.007\pm 0.010$       &      $1.485\pm 0.028$     &        $1.458\pm 0.013$  &     $0.795\pm 0.010$      &     $0.905\pm 0.017$        \\[0.1cm]
\textbf{AGCRN} \cite{bai2020adaptive}&$1.120\pm 0.010$      &    $1.116\pm 0.017$       &   $2.039\pm 0.022$       &    $2.010\pm 0.009$       &      $1.469\pm 0.030$     &        $1.465\pm 0.026$  &     $0.788\pm 0.011$      &     $0.832\pm 0.020$        \\[0.1cm]
\hline
\end{tabular}
}
\end{table*}

\subsection{Predictive Performance}\label{predperform}
The forecasting experiments focus on the evaluation of the recurrent graph neural networks implemented in our framework. We compare the predictive performance under two specific backpropagation regimes which can be used to train these recurrent models:
\begin{itemize}
    \item \textbf{Incremental:} After each temporal snapshot the loss is backpropagated and model weights are updated. This would need as many weight updates as the number of temporal snapshots.
    \item \textbf{Cumulative:} When the loss from every temporal snapshot is aggregated it is backpropagated and weights are updated with the optimizer. This requires one weight update per epoch.
\end{itemize}
\subsubsection{Experimental settings} Using 90\% of the temporal snapshots for training, we evaluated the forecasting performance on the last 10\% by calculating the average mean squared error from 10 experimental runs. We used models with a recurrent graph convolutional layer which had 32 convolutional filters. The spatiotemporal layer was followed by the rectified linear unit \cite{nair2010rectified} activation function and during training time we used a dropout of 0.5 for regularization \cite{srivastava2014dropout} after the spatiotemporal layer. The hidden representations were fed to a fully connected feedforward layer which outputted the predicted scores for each spatial unit. The recurrent models were trained for 100 epochs with the Adam optimizer \cite{kingma_adam_2014} which used a learning rate of $10^{-2}$ to minimize the mean squared error. 
\subsubsection{Experimental findings} Results are presented in Table \ref{tab:predictive_performance} where we also report standard deviations around the test set mean squared error and bold numbers denote the best performing model under each training regime on a dataset. Our experimental findings demonstrate multiple important empirical regularities which have important practical implications. Namely these are the following:
\begin{enumerate}
    \item Most recurrent graph neural networks have a similar predictive performance on these regression tasks. In simple terms there is not a single model which acts as \textit{silver bullet}. This also postulates that the model with the lowest training time is likely to be as good as the slowest one.
    \item Results on the Wikipedia Math dataset imply that a cumulative backpropagation strategy can have a detrimental effect on the predictive performance of a recurrent graph neural network. When computation resources are not a bottleneck an incremental strategy can be significantly better.
\end{enumerate}

\subsection{Runtime Performance}
The evaluation of the PyTorch Geometric Temporal runtime performance focuses on manipulating the input size and measuring the time needed to complete a training epoch. We investigate the runtime under the incremental and cumulative backpropagation strategies. 

\subsubsection{Experimental settings} The runtime evaluation used the GConvGRU model \cite{gconvlstm} with the hyperparameter settings described in Subsection \ref{predperform}. We measured the time needed for doing a single epoch over a sequence of 100 synthetic graphs. Reference Watts-Strogatz graphs in the snapshots of the dynamic graph with temporal signal iterator had binary labels, $2^{10}$ nodes, $2^5$ edges per node and $2^5$ node features. Runtimes were measured on the following hardware:

\begin{itemize}
    \item \textbf{CPU:} The machine used for benchmarking had 8 \textit{Intel 1.00 GHz i5-1035G1} processors. 
    \item \textbf{GPU:} We utilized a machine with a single \textit{Tesla V-100} graphics card for the experiments.
\end{itemize}

\begin{figure}[h!]

	\centering
	\begin{tikzpicture}[scale=0.45,transform shape]
	\tikzset{font={\fontsize{14pt}{12}\selectfont}}
	\begin{groupplot}[group style={group size=2 by 2,
		horizontal sep=40pt, vertical sep=50pt,ylabels at=edge left},
	width=0.54\textwidth,
	height=0.3375\textwidth,
	ymin=0,
	ymax=24,
	legend columns=3,
every tick label/.append style={font=\bf},
    y tick label style={
        /pgf/number format/.cd,
            fixed,
            fixed zerofill,
            precision=0,
        /tikz/.cd
    },
 enlarge x limits=true,
	grid=major,
	grid style={dashed, gray!40},
	scaled ticks=false,
	inner axis line style={-stealth}]

% \nextgroupplot[
%    ybar=0pt,
%      every tick/.style={
%        black,
%        semithick,
%      },
%legend image post style={solid},     
%    bar width=9pt,
%    enlargelimits=0.15,
%    legend style={at={(0.5,-0.15)},
%      anchor=north,legend columns=-1},
%    ylabel={Runtime increase in \%},
%    ytick={0,20,40,60,80,100},
%    xlabel={$\log_2$ Number of nodes},
%    symbolic x coords={6,8,10,12,14},
%    xtick={6,8,10,12,14}]

%\coordinate (BASE) at (axis cs:6,.5);
%\coordinate (O1) at (rel axis cs:0,0);
%\coordinate (O2) at (rel axis cs:1,0);

%\coordinate (SKETCH) at (axis cs:EN,.684);
%\coordinate (O1) at (rel axis cs:0,0);
%\coordinate (O2) at (rel axis cs:1,0);

%\draw [blue,ultra thick,sharp plot,opacity=0.4] (SKETCH -| O1) -- (SKETCH-| O2);

%\coordinate (LINEAR) at (axis cs:EN,.6);
%\coordinate (O1) at (rel axis cs:0,0);
%\coordinate (O2) at (rel axis cs:1,0);

%\draw [red,ultra thick, sharp plot,opacity=0.4] (LINEAR -| O1) -- (LINEAR -| O2);

%\coordinate (NEURAL) at (axis cs:EN,.7);
%\coordinate (O1) at (rel axis cs:0,0);
%\coordinate (O2) at (rel axis cs:1,0);

%\draw [green,ultra thick,sharp plot,opacity=0.4] (NEURAL -| O1) -- (NEURAL -| O2);

%\addplot [fill=blue!35]  coordinates {
%(6,28.631)
%(8,31.968)
%(10,27.312)
%(12,32.025)
%(14,27.578)
%};
%\addplot [fill=blue!55]  coordinates {
%(6,43.726)
%(8,41.253)
%(10,44.325)
%(12,42.603)
%(14,41.587)
%};
%\addplot [fill=blue!75] coordinates {
%(6,59.518)
%(8,59.114)
%(10,54.145)
%(12,62.333)
%(14,57.91)
%};

 \nextgroupplot[
   xlabel=$\log_2$ Number of nodes,
    ybar=0pt,
      every tick/.style={
        black,
        semithick,
      },
    bar width=9pt,
    enlargelimits=0.17,
    ylabel={Runtime in seconds},
    ytick={0,8,16,24},
    legend style={at={(0.5,-0.15)},
      anchor=north,legend columns=-1},
    symbolic x coords={8, 9, 10, 11, 12},
    xtick={8, 9, 10, 11, 12}]

\addplot [fill=red!25]  coordinates {
(8,2.599700164794922)
(9,3.375516200065613)
(10,7.114032745361328)
(11,13.63241662979126)
(12,26.32260463237762)

};
\addplot [fill=red!55]  coordinates {
(8,2.7902231216430664)
(9,4.056894898414612)
(10,6.870487380027771)
(11,11.959950876235961)
(12,23.78959305286408)

};
\addplot [fill=blue!25] coordinates {
(8,0.9580083847045898)
(9,0.9671754837036133)
(10,0.99382164478302)
(11,1.039305567741394)
(12,1.116880178451538)
};

\addplot [fill=blue!55] coordinates {
(8,0.7704073667526246)
(9,0.7901243925094604)
(10,0.8253968715667724)
(11,0.8712318420410157)
(12,0.9239952564239502)
};

% \nextgroupplot[
%   xlabel=$\log_2$ Number of node features,
%    ybar=0pt,
%      every tick/.style={
%        black,
%        semithick,
%      },
%    bar width=9pt,
%    enlargelimits=0.15,
%    legend image post style={solid},
%    legend style={at={(0.5,-0.15)},
%      anchor=north,legend columns=-1},
%    ylabel={Runtime increase in \%},
%    symbolic x coords={5, 6, 7, 8, 9},
%    xtick={5, 6, 7, 8, 9},
%    ytick={0,20,40,60,80,100},
%    	legend style = { column sep = 10pt, legend columns = 1, legend to name = grouplegend}   ]

%\addplot [fill=blue!35]  coordinates {
%(5,24.359)
%(6,25.123)
%(7,25.416)
%(8,25.121)
%(9,26.636)
%};
%\addplot [fill=blue!55]  coordinates {
%(5,41.397)
%(6,40.085)
%(7,42.458)
%(8,41.757)
%(9,42.806)
%};
%\addplot [fill=blue!75] coordinates {
%(5,53.221)
%(6,51.099)
%(7,53.012)
%(8,53.439)
%(9,56.743)
%};

 \nextgroupplot[
   xlabel=$\log_2$ Number of edges per node,
    ybar=0pt,
      every tick/.style={
        black,
        semithick,
      },
    bar width=9pt,
    enlargelimits=0.17,
    legend columns=4,
    legend image post style={solid},
    legend style={at={(0.5,-0.25)},nodes={scale=1.5, transform shape}, 
      anchor=north,legend columns=-1},
yticklabels={,,,,,},
ytick={0,8,14,24},
    symbolic x coords={2,3,4,5,6},
    xtick={2, 3, 4, 5, 6},
    	legend style = { column sep = 10pt, legend columns = 1, legend to name = grouplegend}   ]

\addplot [fill=red!25]  coordinates {
(2,3.121696376800537)
(3,3.722192788124085)
(4,4.736585903167725)
(5,6.65434992313385)
(6,10.79632422924042)
};

\addplot [fill=red!55]  coordinates {
(2,2.5616374731063845)
(3,3.3613226652145385)
(4,3.621216130256653)
(5,6.071313405036927)
(6,10.076012802124023)

};

\addplot [fill=blue!25] coordinates {
(2,0.9738906860351563)
(3,0.9814468860626221)
(4,0.9865357637405395)
(5,1.0063757181167603)
(6,1.0106365919113158)
};
\addplot [fill=blue!55] coordinates {
(2,0.7816596984863281)
(3,0.7865363597869873)
(4,0.7843228578567505)
(5,0.8068660736083985)
(6,0.819727897644043)
};

 \nextgroupplot[
   xlabel=$\log_2$ Number of node features,
    ybar=0pt,
      every tick/.style={
        black,
        semithick,
      },
    bar width=9pt,
    enlargelimits=0.17,
    ylabel={Runtime in seconds},
    ytick={0,8,16,24},
    legend style={at={(0.5,-0.15)},
      anchor=north,legend columns=-1},
    symbolic x coords={3, 4, 5, 6, 7},
    xtick={3, 4, 5, 6, 7}]

\addplot [fill=red!25]  coordinates {
(3,5.222227787971497)
(4,6.702340817451476)
(5,7.59520480632782)
(6,8.935064244270325)
(7,15.80709080696106)

};
\addplot [fill=red!55]  coordinates {
(3,4.640259575843811)
(4,5.733296608924865)
(5,6.613083243370056)
(6,9.351088762283325)
(7,14.654789972305299)
};
\addplot [fill=blue!25] coordinates {
(3,0.9964099645614624)
(4,0.9970572710037231)
(5,1.004029107093811)
(6,1.0235532522201538)
(7,1.0380285739898683)
};

\addplot [fill=blue!55] coordinates {

(3,0.8171007394790649)
(4,0.8103402376174926)
(5,0.8154025554656983)
(6,0.84840989112854)
(7,0.8576318740844726)
};

% \nextgroupplot[
%   xlabel=$\log_2$ Number of node features,
%    ybar=0pt,
%      every tick/.style={
%        black,
%        semithick,
%      },
%    bar width=9pt,
%    enlargelimits=0.15,
%    legend image post style={solid},
%    legend style={at={(0.5,-0.15)},
%      anchor=north,legend columns=-1},
%    ylabel={Runtime increase in \%},
%    symbolic x coords={5, 6, 7, 8, 9},
%    xtick={5, 6, 7, 8, 9},
%    ytick={0,20,40,60,80,100},
%    	legend style = { column sep = 10pt, legend columns = 1, legend to name = grouplegend}   ]

%\addplot [fill=blue!35]  coordinates {
%(5,24.359)
%(6,25.123)
%(7,25.416)
%(8,25.121)
%(9,26.636)
%};
%\addplot [fill=blue!55]  coordinates {
%(5,41.397)
%(6,40.085)
%(7,42.458)
%(8,41.757)
%(9,42.806)
%};
%\addplot [fill=blue!75] coordinates {
%(5,53.221)
%(6,51.099)
%(7,53.012)
%(8,53.439)
%(9,56.743)
%};

 \nextgroupplot[
   xlabel=$\log_2$ Number of filters,
    ybar=0pt,
      every tick/.style={
        black,
        semithick,
      },
    bar width=9pt,
    enlargelimits=0.17,
    legend columns=4,
    legend image post style={solid},
    legend style={at={(0.5,-0.25)},nodes={scale=1.5, transform shape}, 
      anchor=north,legend columns=-1},
yticklabels={,,,,,},
ytick={0,8,16,24},
    symbolic x coords={2, 3, 4, 5, 6},
    xtick={2, 3, 4, 5, 6},
    	legend style = { column sep = 10pt, legend columns = 1, legend to name = grouplegend, font=\small}  ]

\addplot [fill=red!25]  coordinates {
(2,6.145688734054565)
(3,5.599588871002197)
(4,6.755943989753723)
(5,8.197818303108215)
(6,12.243383158683778)
};
\addlegendentry{Incremental CPU}
\addplot [fill=red!55]  coordinates {
(2,6.240883541107178)
(3,5.504344320297241)
(4,5.871164345741272)
(5,7.961208701133728)
(6,11.281559109687805)

};
\addlegendentry{Cumulative CPU}
\addplot [fill=blue!25] coordinates {

(2,1.0071399927139282)
(3,1.01289701461792)
(4,1.0120246410369873)
(5,1.0133137464523316)
(6,1.0780073165893556)};\addlegendentry{Incremental GPU}
\addplot [fill=blue!55] coordinates {

(2,0.8335606336593628)
(3,0.8262794733047485)
(4,0.8225301504135132)
(5,0.8182852745056153)
(6,0.8570517539978028)};
\addlegendentry{Cumulative GPU}

	\end{groupplot}

	\node at ($(group c2r2) + (-4.7cm,-4.4cm)$) {\ref{grouplegend}}; 
	\end{tikzpicture}
	
	\caption{The average time needed for doing an epoch on a dynamic graph -- temporal signal iterator of Watts Strogatz graphs with a recurrent graph convolutional model.}\label{fig:runtime}
\end{figure}
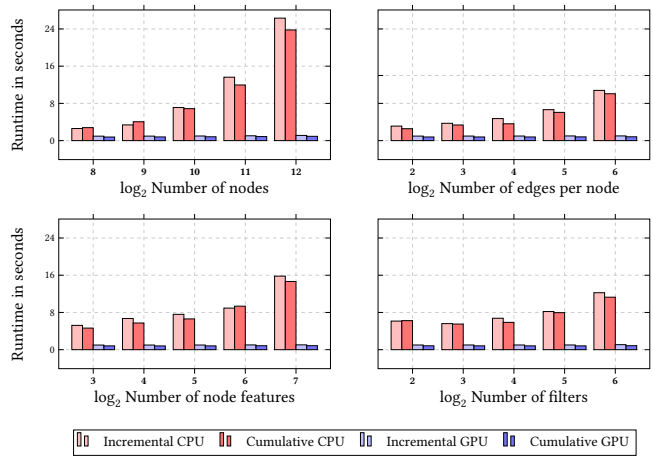

\subsubsection{Experimental findings} We plotted the average runtime calculated from 10 experimental runs on Figure \ref{fig:runtime} for each input size. Our results about runtime have two important implications about the practical application of our framework:
\begin{enumerate}
    \item The use of a cumulative backpropagation strategy only results in marginal computation gains compared to the incremental one.
    \item On temporal sequences of large dynamically changing graphs the GPU aided training can reduce the time needed to do an epoch by a whole magnitude.
\end{enumerate}

\section{Conclusions and Future Directions}\label{sec:conclusions}
In this paper we discussed \textit{PyTorch Geometric Temporal} the first deep learning library designed for neural spatiotemporal signal processing. We reviewed the existing geometric deep learning and machine learning techniques implemented in the framework. We gave an overview of the general machine learning framework design principles, the newly introduced input and output data structures, long-term project viability and discussed a case study with source-code which utilized the library. Our empirical evaluation focused on (a) the predictive performance of the models available in the library on real world datasets which we released with the framework; (b) the scalability of the methods under various input sizes and structures.

Our work could be extended and it also opens up opportunities for novel geometric deep learning and applied machine learning research. A possible direction to extend our work would be the consideration of continuous time  or time differences between temporal snapshots which are not constant. Another opportunity is the inclusion of temporal models which operate on curved spaces such as hyperbolic and spherical spaces. We are particularly interested in how the spatiotemporal deep learning techniques in the framework can be deployed and used for solving high-impact practical machine learning tasks. 
\bibliographystyle{ACM-Reference-Format}
\bibliography{main}
\end{document}